\pdfoutput=1
\documentclass[11pt]{article}

\usepackage[]{acl}
\usepackage{times}
\usepackage{latexsym}

\usepackage[T1]{fontenc}
\usepackage[utf8]{inputenc}

\usepackage{microtype}

\usepackage{inconsolata}
\usepackage{colortbl}
\usepackage{diagbox}

\usepackage{booktabs} 
\usepackage{graphicx} 
\usepackage{CJKutf8}
\usepackage{amsmath} 
\usepackage[textwidth=2cm]{todonotes}
\usepackage{subcaption}
\usepackage{tabularx}
\usepackage{adjustbox}
\usepackage{multirow}
\usepackage{verbatim} % For the verbatim environment
\usepackage{mdframed}
\usepackage{longfbox}
\usepackage{xcolor}
\usepackage{tikz}
\definecolor{darkpastelred}{rgb}{0.76, 0.23, 0.13}

\newcommand{\mybox}[1]{\tikz[baseline=(MeNode.base)]{\node[rounded corners, fill=gray!20](MeNode){#1};}}

\title{Do We Need Language-Specific Fact-Checking Models? \\ The Case of Chinese}

\author{
Caiqi Zhang$^{1}$, 
Zhijiang Guo$^{2}$, 
Andreas Vlachos$^{2}$
\\
$^1$Language Technology Lab, University of Cambridge \\
$^2$Department of Computer Science and Technology, University of Cambridge \\
\texttt{\{cz391, zg283, av308\}@cam.ac.uk}}

\newcommand{\eg}{\textit{e}.\textit{g}.,\ }

\newcommand{\lpm}[2]{$#1 \pm \scriptstyle{#2}$}
\newcommand{\blpm}[2]{$\mathbf{#1 \pm \scriptstyle{#2}}$}

\begin{document}
\maketitle

\newpage
\begin{abstract}

This paper investigates the potential benefits of language-specific fact-checking models, focusing on the case of Chinese using CHEF dataset. To better reflect real-world fact-checking, we first develop a novel Chinese document-level evidence retriever, achieving state-of-the-art performance. We then demonstrate the limitations of translation-based methods and multilingual language models, highlighting the need for language-specific systems. To better analyze token-level biases in different systems, we construct an adversarial dataset based on the CHEF dataset, where each instance has a large word overlap with the original one but holds the opposite veracity label. Experimental results on the CHEF dataset and our adversarial dataset show that our proposed method outperforms translation-based methods and multilingual language models and is more robust toward biases, emphasizing the importance of language-specific fact-checking systems.\footnote{\url{https://github.com/caiqizh/FC_Chinese}}

% \footnote{Our dataset and code will be publicly available.}.
% 
\end{abstract}

% We further point out the cultural biases in CHEF and create an adversarial dataset. Experiments on our newly proposed dataset show a significant decrease in both accuracy and F1 score. Overall, our study highlights the necessity of devising language-specific fact-checking models. 
\section{Introduction}
%\vspace{-2mm}
Since manual fact-checking requires significant time and resources, there has been growing interest in automated fact-checking in recent years~\citep{2018graves,nakov2021}.
While misinformation exists in various languages, studies have predominantly focused on claims and evidence in English~\citep{GuoSV22,mubashara2023multimodal}. Current research in fact-checking in other languages often lacks grounding in real-world claims \citep{chang-etal-2023-xfever} or is constrained to a single domain, like COVID-19 \citep{shahifakecovid}. 

In this paper, we raise the question: \textit{Should we develop language-specific fact-checking models, or can we effectively utilize existing English models by translating claims and evidence into English?}
We present a case study focused on Mandarin Chinese to investigate it for two reasons. Firstly, Chinese is widely spoken by over a billion people and possesses unique linguistic characteristics different from English~\citep{Yang2017AnIO, fei2023differences}.
Secondly, Chinese is the only language other than English that has a real-world evidence-based dataset annotated manually, \textit{i.e.,} CHEF~\citep{hu-etal-2022-chef}). %In contrast, datasets in other languages either lack evidence annotations or real-world claims. For example, 
This is in contrast to other popular datasets such as X-FACT \citep{gupta-srikumar-2021-x}, which depend on Google search results without evidence annotation, and DanFEVER \citep{norregaard-derczynski-2021-danfever}, a Danish translation of FEVER, which do not address real-life fact-checking needs.

% Proper evidence is essential for systems to validate claims accurately, avoiding dependence on superficial biases \citep{jiang-etal-2020-hover, Borel+2023}. 

\begin{CJK*}{UTF8}{gbsn}
\begin{table}[t!]
    \small
    \centering
    \begin{tabular}{p{0.99\columnwidth}}
    \toprule
    Original: 广东两名小学生\textcolor{blue}{提干}，引发大量讨论。\\
    Translated: Two primary school students in Guangdong \textcolor{red}{raised eyebrows} (\textcolor{blue}{were promoted}), sparking discussion. \\
    
    \textbf{ChatGPT:} \mybox{\textsc{refuted}} \textbf{CHEF Label:} \mybox{\textsc{supported}} \\
    \midrule
    Claim 1: 中国超八成地下水遭受污染，不能饮用。\\
    (\textit{Over 80\% of China's groundwater is polluted and is unfit for drinking.}) \\
    Claim 2: 中国高铁辐射严重引发不孕。(\textit{Radiation from China's high-speed rail seriously causes infertility.}) \\
    \textbf{ChatGPT:} \mybox{\textsc{supported}} \textbf{CHEF Label:} \mybox{\textsc{refuted}}
    \\
    \bottomrule
    \end{tabular}
    %\vspace{-2mm}
    \caption{\small Upper section: the challenge in accurate translation (\textcolor{red}{Red}: Incorrect, \textcolor{blue}{Blue}: Correct); Lower section: the bias of multilingual LLMs towards certain claims.}
    \label{tab:example}
    \vspace{-3mm}
\end{table}
\end{CJK*}

% \begin{table*}[ht!]
%     \centering
%     %\scalebox{0.8}{
%     \begin{footnotesize}
% \begin{tabular}{l c c c c c c}
%     \toprule
%     \multirow{2}{*}{\textbf{Verifiers}} & \multicolumn{2}{c}{Semantic Ranker} & \multicolumn{2}{c}{Document-level Retriever} & \multicolumn{2}{c}{Gold Evidence} \\
%     \cmidrule(lr){2-3} \cmidrule(lr){4-5} \cmidrule(lr){6-7}
%      & Accuracy & Macro F1 & Accuracy & Macro F1 & Accuracy & Macro F1 \\
%     \midrule
%     BERT-base  & 63.00 & 62.88 & 67.66 & 67.66 & 77.79 & 77.62 \\ 
%     Attention-based & 64.01 & 63.65 & 69.00 & 68.35 & 78.56 & 78.46 \\ 
%     Graph-based & 62.43 & 62.42 & 69.25 & 69.14 & 78.95 & 78.39 \\ 
%     RoBERTa-large & 66.37 & 66.24 & 72.31 & 72.31 & 79.38 & 79.47 \\
%     \midrule
%     GPT-3.5-Turbo & 53.29 & 51.46 & 55.45 & 51.32 & 58.79 & 54.97 \\
%     GPT-4-Turbo & 65.78 & 62.35 & 69.17 & 69.01 & 73.67 & 73.96 \\
%     DeBERTa-large & \textbf{69.89} & \textbf{68.34} & \textbf{74.50} & \textbf{74.46} & \textbf{81.46} & \textbf{81.15} \\ 
%     \midrule
%     Google-Translate & 59.23 & 59.76 & 60.15 & 61.29 & 66.84 & 66.57 \\
%     GPT-4-Translate & 62.17 & 62.43 & 62.36 & 60.01 & 67.95 & 67.82 \\
%     \bottomrule
%     \end{tabular}
%     \end{footnotesize}    
%     \caption{Results on CHEF. The translated baseline uses our document-level retriever and DeBERTa-large claim verifier, translated via Google Translator and GPT-4.}
%     \label{tab:pipeline_results}
%     \vspace{-4mm}
% \end{table*}

\begin{table*}[ht!]
    \centering
    %\scalebox{0.8}{
    \begin{footnotesize}
\begin{tabular}{c l c c c c c c}
    \toprule
    & \multirow{2}{*}{\textbf{\diagbox[innerwidth=2.5cm,height=9mm]{Verifiers}{Retrievers}}} & \multicolumn{2}{c}{Semantic Ranker} & \multicolumn{2}{c}{DLR} & \multicolumn{2}{c}{Gold Evidence} \\
    \cmidrule(lr){3-4} \cmidrule(lr){5-6} \cmidrule(lr){7-8}
    &  & Acc. & Macro F1 & Acc. & Macro F1 & Acc. & Macro F1 \\
     \midrule
   \multirow{2}{*}{Translation} &  GT+DeBERTa  & 59.23 & 59.76 & 60.15 & 61.29 & 66.84 & 66.57 \\
    & GPT-4+DeBERTa & 62.17 & 62.43 & 62.36 & 60.01 & 67.95 & 67.82 \\
    \midrule
   \multirow{4}{*}{Multilingual LM} & mDeBERTa \citep{he2021debertav3} & 65.34	&64.79	&71.89&	70.97	&79.86	&79.63\\
   & GPT-4 + mDeBERTa & 55.24	&54.38	&56.78&	58.69&	60.31&	61.75 \\
   & GPT-3.5-Turbo \citep{gpt3.5} & 53.29 & 51.46 & 55.45 & 51.32 & 58.79 & 54.97 \\
    & GPT-4-Turbo \cite{gpt4}& 65.78 & 62.35 & 69.17 & 69.01 & 73.67 & 73.96 \\
    \midrule
  \multirow{5}{*}{Chinese Specific} &  BERT-base \citep{devlin-etal-2019-bert} & 63.00 & 62.88 & 67.66 & 67.66 & 77.79 & 77.62 \\ 
   &  Attention-based \citep{gupta-srikumar-2021-x} & 64.01 & 63.65 & 69.00 & 68.35 & 78.56 & 78.46 \\ 
    & Graph-based \citep{liu-etal-2020-fine}& 62.43 & 62.42 & 69.25 & 69.14 & 78.95 & 78.39 \\ 
    &  Chinese RoBERTa \citep{cui-etal-2020-revisiting}& 66.37 & 66.24 & 72.31 & 72.31 & 79.38 & 79.47 \\
   &  Chinese DeBERTa \citep{fengshenbang}& \textbf{69.89} & \textbf{68.34} & \textbf{74.50} & \textbf{74.46} & \textbf{81.46} & \textbf{81.15} \\ 
    \bottomrule
    \end{tabular}
    \end{footnotesize}    
    %\vspace{-2mm}
    \caption{Results on CHEF. All the DeBERTa and RoBERTa models are of \textit{-large} version.}
    \label{tab:pipeline_results}
    \vspace{-2mm}
\end{table*}

%To better align our work with real-world fact-checking,
To study the question proposed,
we first develop a novel Chinese document-level evidence retriever (DLR), which outperforms state-of-the-art models by 10\% in terms of accuracy and Macro F1 in CHEF. Paired with either DLR-retrieved or annotated gold evidence, we then demonstrate the limitations of translation-based methods (\textit{i.e.,} first translating Chinese claims and evidence into English and then applying English fact-checking models on translated data) or multilingual language models in fact-checking. We further identify the cultural biases in CHEF and create an adversarial dataset. Experiments on our newly proposed dataset show a significant decrease in both accuracy and F1 score. Overall, our study highlights the necessity of devising language-specific fact-checking models. 
% \section{Language-Specific Models or Others?}
%\section{Experiments on CHEF}
\section{Language-specific models vs translation and multilingual models}

%\vspace{-2mm}
To construct a Chinese fact-checking system, two straightforward approaches are direct translation from Chinese to English and the application of multilingual LLMs. However, as demonstrated in Table \ref{tab:example}, translation from Chinese to English may result in inaccuracies, particularly with idiomatic expressions or language-specific phrases \citep{shao-etal-2018-evaluating}. Furthermore, LLMs like ChatGPT, predominantly trained on English data \citep{lai-etal-2023-chatgpt,Hu2023DoLL}, tend to reflect Western norms, values and biases \citep{naous2023having, masoud2023cultural, wang2023countries}, making them less effective for fact-checking in other languages. Moreover, LLMs also suffer from hallucination problem \citep{zhang2023sirens}. Table \ref{tab:example} shows examples of scientifically refuted claims that GPTs incorrectly accept, alongside the evidence retrieved.
To examine the abovementioned limitations in a systematic way, we conduct experiments on a large scale Chinese evidence-based dataset, CHEF. %\cite{hu-etal-2022-chef}.
%\citep{nie2019combining, liu-etal-2020-fine}. Similar to English fact-checking systems~\citep{thorne2018}, evidence retrieval also plays a crucial role in Chinese fact-checking, making the system more interpretable and reliable~\citep{samarinas-etal-2021-improving, bekoulis-etal-2021-understanding}. Furthermore, evidence retrieval is the bottleneck of a fact-checking system as the accuracy of veracity prediction heavily relies on the quality of the evidence~\citep{hu-etal-2022-chef}. 

\noindent
\paragraph{Retrievers} 
Evidence retrieval is crucial in the fact-checking pipeline \citep{augenstein2019multifc, jiang-etal-2020-hover}. To align our work with real-world fact-checking, we develop a state-of-the-art Chinese evidence retrieval component. Our novel Document-level Retriever (DLR) enhances evidence retrieval by considering the context of evidence sentences, unlike prior approaches that isolate evidence selection to pairwise sentence classification \citep{hu-etal-2022-chef}. Inspired by~\citet{stammbach-2021-evidence}, 
we train a retriever to assign a score to each Chinese token within an evidence document and then aggregate these token scores at the sentence level. In particular, we fine-tune a BigBird~\citep{zaheer2020big} to assign a value of 1 to tokens that belong to annotated evidence for a claim, while assigning a value of 0 to all other tokens. During inference, we compute the average scores for all tokens within each sentence. If the resulting average score exceeds 0.5, we classify the sentence as evidence. We compare our proposed document-level retriever with the annotated gold evidence, and Semantic Ranker~\citep{nie2019combining, liu-etal-2020-fine} used in CHEF. To eliminate the effects of evidence retrieval, we also utilize the gold evidence for the ideal case. More training details of DLR can be found in Appendix \ref{sec:dlr}.

\paragraph{Verifiers} For the verifiers in Table \ref{tab:pipeline_results}, we mainly include the baselines in \citet{hu-etal-2022-chef} with some of our variations: (1) For the translation models, we first translate the evidence and claims via Google Translator (GT) and GPT-4, then apply the DeBERTa-large verifier. (2) The cross-lingual approach \textit{GPT-4 + mDeBERTa} involves training multilingual DeBERTa-large on data translated from Chinese to English, followed by evaluation of the original CHEF. For the GPT models,\footnote{Our experiments are conducted in Feb 2024.} we use 5 shots for in-context learning. Other models are all trained and tested on original CHEF. We provide detailed experimental settings in the Appendix \ref{exp_app}. 

\begin{CJK*}{UTF8}{gbsn}

\begin{table*}[ht!]
    \fontsize{7}{7}\selectfont
    \renewcommand{\arraystretch}{1.5}
\begin{minipage}{.5\linewidth}
\centering
\begin{tabular}{p{4.5cm}ll}
    \toprule
        Word & LMI($10^{-6}$) & $p(l|w)$ \\ 
        \midrule
        中国 \ (China) & 1189 & 0.56 \\ 
        电影 \  (Movie) & 1008 & 0.84 \\ 
        国际  \ (International) & 629 & 0.80 \\ 
        发布 \  (Release/Announce) & 599 & 0.74 \\ 
        金融  \ (Finance) & 593 & 0.66 \\ 
        亿元 \  (Hundred Million Yuan) & 500 & 0.66 \\ 
        外交 \  (Diplomacy/Foreign Affairs) & 496 & 0.85 \\ 
        外交部 \  (Ministry of Foreign Affairs) & 481 & 0.92 \\ 
        人民币 \  (RMB/Chinese Yuan) & 469 & 0.84 \\ 
        银行  \ (Bank) & 469 & 0.63 \\ 
    \bottomrule
    \end{tabular}
\end{minipage}
\begin{minipage}{.5\linewidth}
\centering
\begin{tabular}{p{4cm}ll}
    \toprule
        Word & LMI($10^{-6}$) & $p(l|w)$ \\ 
        \midrule
        病毒 \  (Virus) & 1105 & 0.66 \\ 
        疫苗 \  (Vaccine) & 1013 & 0.64 \\ 
        台湾 \  (Taiwan) & 962 & 0.77 \\ 
        可以 \  (Can/Be able to) & 901 & 0.72 \\ 
        出现 \  (Appear) & 478 & 0.74 \\ 
        肺炎 \  (Pneumonia) & 475 & 0.70 \\ 
        手机 \  (Mobile phone) & 451 & 0.77 \\ 
        冠状 \  (Coronary) & 414 & 0.93 \\ 
        日本 \  (Japan) & 402 & 0.72 \\ 
        感染 \  (Infection) & 395 & 0.66 \\ 
        \bottomrule
    \end{tabular}
\end{minipage}
\vspace{-2mm}
\caption{Top 10 LMI-ranked phrases in the train set of CHEF for \textsc{supported} (left) and \textsc{refuted} (right).}% with $p(l|w)$.}
\label{tab:top20}
\vspace{-2mm}
\end{table*}
\end{CJK*}

\paragraph{Results on CHEF} As shown in Table \ref{tab:pipeline_results}, our system that combines DLR and \textit{Chinese DeBERTa}, yields the best results with an accuracy of 74.50\% and a Macro F1 score of 74.46\%. There is an improvement of over 10\% compared to the best translation-based result (\textit{GPT-4+DeBERTa}). 
For the multilingual LMs, the cross-lingual approach (\textit{GPT-4+mDeBERTa}) show poor performance. This is probably because of the disparity between the languages of the training and testing sets. On the contrary, \textit{mDeBERTa}, trained on Chinese CHEF, achieves competitive performance on par with \textit{Chinese RoBERTa} and outperforming several baseline models. Overall, the results suggest that training a model specifically on Chinese, whether it be a multilingual or monolingual model, is more beneficial than relying on English-centric ones.

\paragraph{Evidence Retrieval} The DLR, paired with different verifiers, improves accuracy and Macro F1 by about 5\% over the Semantic Ranker. Regarding the recall of human-annotated gold evidence, DLR leads to 10\% higher Recall@5 (Table~\ref{tab:retriever_overall}). We also find that our new retriever can retrieve evidence pieces which, when considered individually cannot verify the claim but, when combined they can. Table~\ref{tab:documet_retriever_example} gives a detailed example in the Appendix~\ref{compare_retrievers}. 

% \noindent
% \textbf{Claim Verification} The pipeline's performance is improved by incorporating RoBERTa and DeBERTa as claim verifiers. The DeBERTa-large yields a notable enhancement, with a 5\% uplift in both accuracy and Macro F1 scores over the best-reported baseline with attention-based retriever and document-level verifier. 

\section{Biases in CHEF}\label{sec:bias_in_chef}
%\vspace{-2mm}
To explore the reasons behind the deficiency of translation services and multilingual LMs, we investigate the biases present in the CHEF dataset in this section. Prior research has demonstrated that fact-checking datasets, such as FEVER~\citep{thorne2018} and MultiFC~\citep{augenstein-etal-2019-multifc}, result in training models that rely on heuristics such as surface-level patterns within claims, potentially impeding their ability to generalize effectively~\citep{schuster-etal-2019-towards, thorne2019}. In this section, we show that while biases are present as in the English language datasets and models, \textbf{they are specific to the Chinese culture}. 

\paragraph{Domain Bias} First, in CHEF, claims are categorized into domains such as politics, society, health, and culture and we find a significant skew in the distribution:\textit{ 64\% of social and 66\% of health claims are \textsc{refuted}}, \textit{while 55\% in politics and 72\% in culture are \textsc{supported}}. Notably, there is an imbalance in the proportion of social and health claims, which collectively constitute 68\% of the total. Figure~\ref{fig:distribution} in the Appendix details the label distribution across domains.

\paragraph{Cultural Bias}  We then examine the correlation between phrases within the claims and the corresponding labels. The word distribution within the training set is analyzed for this purpose. Initially, all claims in the training set are tokenized by Chinese text segmentation tool, jieba.\footnote{https://github.com/fxsjy/jieba} 
The average length of the words is 2.39 characters. Then, two metrics are employed to assess the correlation between phrases and labels. Following~\citet{schuster-etal-2019-towards},  first we use $p(l|w)$ to calculate the probability of a label $l$ given the presence of a specific phrase $w$ in the claim. As this metric tends to exhibit bias towards low-frequency words, the second metric utilizes Local Mutual Information (LMI;~\citealt{evert2005statistics}) to identify high-frequency $n$-grams that display a strong correlation with a particular label. The $p(l|w)$ and LMI between phrase $w$ and label $l$ is defined as follows:
%\vspace{-\baselineskip}
\begin{align}
    \fontsize{9}{11}\selectfont p(l \mid w) &= \frac{\operatorname{count}(w, l)}{\operatorname{count}(w)}  \\
    \fontsize{9}{11}\selectfont L M I(w, l)&=p(w, l) \cdot \log \left(\frac{p(l \mid w)}{p(l)}\right)
\end{align}
%\vspace{-\baselineskip}
%\normalsize
where we follow~\citet{schuster-etal-2019-towards} to estimate $p(l)$ by $\frac{\operatorname{count}(l)}{|D|}, p(w, l)$ by $\frac{\operatorname{count}(w, l)}{|D|}$ and $|D|$ is the number of occurrences of all $n$-grams. % in the dataset.

\begin{table*}[ht!]
    \centering
    \begin{footnotesize}
    \begin{tabular}{lcccccc}
    \toprule
    & \multicolumn{2}{c}{Original 250 pairs} & \multicolumn{2}{c}{Generated 750 pairs} & \multicolumn{2}{c}{Full 1000 pairs} \\
    \cmidrule(lr){2-3} \cmidrule(lr){4-5} \cmidrule(lr){6-7}
    & Accuracy & F1 Score & Accuracy & F1 Score & Accuracy & F1 Score \\
    \midrule
    GPT-4 + DeBERTa &77.34&	75.26&	43.68&	44.57&	52.10 & 53.04 \\
    \midrule
    mDeBERTa &82.45&	80.68&	53.27&	51.09&	60.56 & 59.68\\
    GPT-3.5-Turbo & 80.00 & 55.25 & 53.73 & 36.78 & 60.30 & 41.39 \\
    \midrule
    BERT-base& 76.35 & 75.36 & 38.56 & 37.62 & 49.06 & 48.72 \\ 
    Attention-based & 78.96 & 78.12 & 39.98 & 39.62 & 51.01 & 49.65 \\
    Graph-based & 79.55 & 76.97 & 39.61 & 38.67 & 49.59 & 49.43 \\
    Chinese DeBERTa & \textbf{86.69} & \textbf{84.98} & \textbf{57.84} & \textbf{54.31} & \textbf{65.01} & \textbf{63.74} \\ 
    \midrule
    GPT-4-Turbo & 85.60 & 60.70 & 65.20 & 47.12 & 70.30 & 50.73 \\
    \bottomrule
    \end{tabular}
    \end{footnotesize}  
        %\vspace{-2mm}
    \caption{Performance comparison of models on the adversarial dataset.  The ``original 250 pairs'' refers to pairs directly extracted from CHEF, while ``generated 750 pairs'' denotes pairs generated using GPT-4.}
    \label{tab:adversarial_results}
    \vspace{-2mm}
\end{table*}

% \begin{table*}[ht!]
%     \centering
%     \begin{footnotesize}
%     \begin{tabular}{lcccccc}
%     \toprule
%     & \multicolumn{2}{c}{Original 250 pairs} & \multicolumn{2}{c}{Generated 750 pairs} & \multicolumn{2}{c}{Full 1000 pairs} \\
%     \cmidrule(lr){2-3} \cmidrule(lr){4-5} \cmidrule(lr){6-7}
%     & Accuracy & F1 Score & Accuracy & F1 Score & Accuracy & F1 Score \\
%     \midrule
%     BERT-base& 76.35 & 75.36 & 38.56 & 37.62 & 49.06 & 48.72 \\ 
%     Attention-based model & 78.96 & 78.12 & 39.98 & 39.62 & 51.01 & 49.65 \\
%     Graph-based model & 79.55 & 76.97 & 39.61 & 38.67 & 49.59 & 49.43 \\
%     GPT-3.5-Turbo & 80.00 & 55.25 & 53.73 & 36.78 & 60.30 & 41.39 \\
%     GPT-4-Turbo & 82.60 & 60.70 & \textbf{65.20} & 47.12 & \textbf{65.30} & 50.73 \\
%     DeBERTa-large & \textbf{86.25} & \textbf{85.78} & 55.01 & \textbf{53.03} & 62.45 & \textbf{62.25} \\ 
%     \bottomrule
%     \end{tabular}
%         \end{footnotesize}    
%     \caption{Performance comparison of models on the adversarial dataset.  The ``original 250 pairs" refers to pairs directly extracted from CHEF, while ``generated 750 pairs" denotes pairs generated using GPT-4.}
%     \label{tab:adversarial_results}
%     \vspace{-3mm}
% \end{table*}

\begin{CJK*}{UTF8}{gbsn}

Table~\ref{tab:top20} lists the top 10 LMI-ranked phrases in the train set of CHEF for \textsc{supported} and \textsc{refuted}. 
We find that while previous research \citep{schuster-etal-2019-towards} found that negation phrasings strongly correspond to \textsc{refuted} labels in FEVER, we did not observe the same pattern in CHEF. 
% Although certain phrases, such as 不能 (cannot), 不要 (do not), and 没有 (none), still appear in the top 10 list, they are not among the highest-ranked phrases.

Prior studies in English datasets, such as Constraint \citep{patwa2020fighting}, have also demonstrated a strong correlation between politician names (\eg Barack Obama and Donald Trump) and refuted claims, however, our research identifies a distinct cultural bias within CHEF. %, setting it apart from these English datasets. 
In CHEF, claims about biomedical and health issues frequently exhibit a strong association with negative labels. Terms such as 病毒\ (virus), 疫苗\ (vaccine), 致癌\ (carcinogenic), and 冠状病毒\ (coronavirus) are more commonly encountered in refuted claims. Conversely, financial terms like 金融\ (finance), 人民币\ (RMB), and 央行\ (People's Bank of China), as well as political terms such as 中国\ (China), 外交部\ (Ministry of Foreign Affairs), tend to carry positive labels. 

One possible reason behind this is that fact-checking in China tends to avoid criticism of hardcore public issues, such as politics, economics, and other current affairs \citep{letscheck}. On the contrary, it focuses more on providing references for everyday decision-making, such as in health. Another political reason could be that the Cyberspace Administration of China keeps a close watch on online news services \citep{letscheck}. Non-state enterprises are not permitted to criticize politics, economics, and other current affairs. Private companies are only authorized to distribute and curate news produced by state-owned media. 
%\citep{letscheck}. 
Furthermore, in CHEF, certain regions such as 台湾\ (Taiwan), 日本\ (Japan), and 美国\ (United States) are commonly associated with the \textsc{refuted} label. This may also reflect the contentious nature of international relations within the realm of Chinese fact-checking.

\end{CJK*}

\section{Adversarial Dataset Construction}
%\vspace{-1mm}
Our analysis revealed the presence of labels and cultural biases specific to the Chinese context (\S~\ref{sec:bias_in_chef}). These biases can significantly impact the performance and fairness of fact-checking models when applied to Chinese-language claims.

We therefore introduce an adversarial dataset derived from the CHEF dataset for a better evaluation of the models. Inspired by \citet{schuster-etal-2019-towards}, to create it %We devise a Symmetric Test Set, 
we pair each claim-evidence instance with a synthetic counterpart where claim and evidence have high word overlap with the original ones but the opposite veracity label (Figure~\ref{fig:adversarial_example}). Under this setting, determining veracity from the claim alone would be equivalent to a random guess. 
%Deviating from labor-intensive manual approaches
Instead of involving human annotators, we opt for the utilization of GPT-4 to generate the dataset. To control the quality, we invited two Chinese native speakers to annotate randomly sampled 25\% of claim-evidence pairs with \textsc{supported}, \textsc{refuted} or \textsc{not enough info}. The results demonstrated strong agreement between humans and GPT-4. They agreed with the dataset labels in 89\% of cases, with a Cohen $\kappa$ of 0.80 \citep{cohen1960coefficient}.
% Specifically, the question posed to the annotators is: ``Do the generated claim and evidence satisfy the symmetric setting?'' 
Our approach overcomes labor-intensive manual annotation and rigid rule-based generation, advocating for automated sample generation using LLMs. This new test set nullifies the benefit of relying exclusively on cues from claims. Details of the dataset construction and the prompt we use can be found in Appendix \ref{dataset_construction}.

\section{Experiments on Adversarial CHEF}
%\vspace{-1mm}
\noindent
\textbf{Results on Adversarial CHEF} Table~\ref{tab:adversarial_results} compares model performance on adversarial versus original data from CHEF.\footnote{Note that here the ``original 250 pairs" use gold evidence, and a binary classification approach, contributing to higher performance metrics than in Table \ref{tab:pipeline_results}.} Since our adversarial dataset was constructed using GPT-4, including it as a verifier will lead to a risk for data leakage, resulting in biased comparisons. Therefore, we have excluded GPT-4 from the evaluation results and have provided its performance metrics for reference only.

All models perform worse on adversarial examples compared to the original CHEF. Specifically, we highlight the following findings: (1) Chinese DeBERTa drops from 86.69\% accuracy on original pairs to 57.84\% and 65.01\% on adversarial subsets. Baselines similarly see over 37\% decreases in both accuracy and F1 scores. This underscores the models' reliance on surface features and reveals the aforementioned biases. 
(2) Compare the translation-based \textit{GPT-4+DeBERTa} (F1 53.04\%), the multilingual \textit{mDeBERTa} (F1 60.56\%), and the \textit{Chinese DeBERTa} (F1 63.74\%), we observe a performance increase with more language-specific models, highlighting the benefits of incorporating Chinese data during pre-training. (3) \textit{GPT-3.5-Turbo}, which has not been fine-tuned on the CHEF dataset, demonstrates better robustness but still underperforms compared to the language-specific \textit{Chinese DeBERTa}. Overall, we recommend future research to use both the original and our adversarial CHEF dataset for a comprehensive evaluation.

\noindent
\textbf{Exposing adversarial examples to the models can improve robustness.} DeBERTa's performance declines less than that of the baselines including BERT, Attention, and Graph-based models when faced with adversarial examples, about 30\% compared to over 37\%, suggesting a higher sensitivity to evidence changes. To investigate the reasons behind the decrease in the model's performance and improve model's robustness, we employ the inoculation fine-tuning method \citep{liu-etal-2019-inoculation}. Results show the performance decline observed in the baselines primarily stems from \textbf{inherent weaknesses} within the model family. In contrast, for the DeBERTa model, gradually exposing it to more adversarial samples leads to a gradual reduction in the performance gap. Inoculation results by fine-tuning the model with different sizes of adversarial examples are provided in Figure~\ref{fig:inoculation_results} in the Appendix~\ref{inoculation}. 
\section{Conclusion}
%\vspace{-1mm}

Our study reveals the shortcomings of English-centric fact-checking systems when applied to Chinese claims, highlighting the failure of translation-based methods due to linguistic and cultural nuances. We introduce a novel system that achieves best-reported results on CHEF and provides an adversarial dataset for continued research, underscoring the need for specialized fact-checking models. 

\section*{Limitations}

% The performance of our document-level retriever, although enhanced compared to the semantic ranker, is still characterized by a relatively low recall rate. This highlights the persisting challenges in evidence retrieval that require further attention and refinement. Another limitation of our study is the availability of evidence-based fact-checking datasets. We could only conduct our analysis on English- and Chinese-language datasets due to the limited availability of evidence-based datasets in other languages. Consequently, more experiments should be conducted to demonstrate the general applicability of our conclusions. 

This study has several notable limitations. Firstly, the performance of our document-level retriever, despite showing improvements over the semantic ranker, still exhibits a relatively low recall rate. This underscores the ongoing challenges in evidence retrieval, which necessitate further research and refinement to enhance the accuracy and reliability of the system.

Secondly, the scope of our analysis is limited to English- and Chinese-language datasets. This is due to the scarcity of real-world, evidence-based fact-checking datasets annotated by humans in other languages. The inclusion of the Chinese dataset in this study is not arbitrary, rather, it's based on two main factors. First, there is a lack of suitable datasets in other languages. Second, several authors of this study are native Chinese speakers, which gives us a distinct advantage in terms of understanding and appreciating the linguistic features and nuances of the Chinese language. This expertise has allowed us to conduct a more thorough and in-depth case study.

Moving forward, we are actively seeking datasets in other languages that align with our research requirements. One potential approach is to construct new datasets based on the recently established fact-checking website, Elections24Check\footnote{\url{https://elections24.efcsn.com/}}. This platform gathers and categorizes fact-checked information for European countries and citizens ahead of the 2024 European Elections. It offers a rich source of political fact-checks, disinformation debunks, prebunking articles, and narrative reports in various European languages, which could be instrumental in expanding the breadth of our research.

\section*{Ethics Statement}
The CHEF dataset employed in our research is accessible to the scientific community, and its use in our experiments presents no conflict of interest. Although the adversarial dataset used in this study was developed with a GPT-4 model, to ensure its integrity and safety, we conducted an extensive manual review to eliminate sensitive or potentially harmful information. This review received approval from our institution's ethics committee. Furthermore, the hourly salary for annotators surpassed the national minimum wage, and all annotators consented to the use of the data.

\section*{Acknowledgements}
Andreas Vlachos is supported by the ERC grant AVeriTeC (GA 865958). Caiqi Zhang is funded by Amazon Studentship. We thank Yulong Chen for his great support on this work.  
%Acknowledgements: Yulong

\bibliography{anthology,custom}

\appendix

\clearpage
\section{Document-level Retriever (DLR)} \label{sec:dlr}

\begin{figure*}
    \centering
    \includegraphics[width=0.80\textwidth]{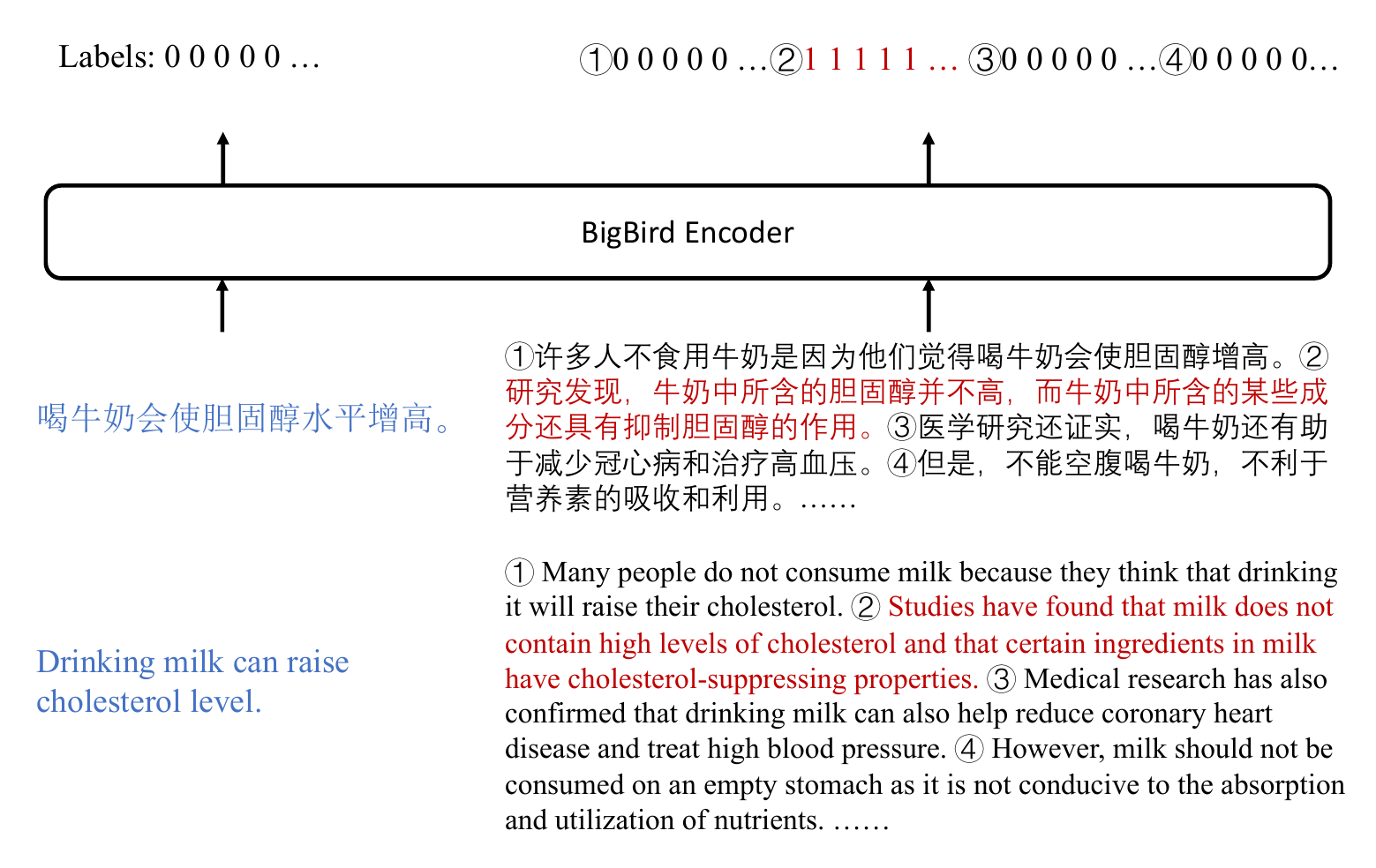}
    \caption{Framework illustration highlighting the usage of BigBird in our approach for evidence sentence retrieval. The claim is represented in \textcolor{blue}{blue}, while the evidence sentence is highlighted in \textcolor{darkpastelred}{red}. }
    \label{fig:bigbird_example}
\end{figure*}

In our approach, we recognize the significance of context in determining whether a given sentence can be considered as evidence. To leverage this contextual information, we assign a score to each token within a document and then aggregate these token scores at the sentence level. Subsequently, we fine-tune a transformer model to assign a value of 1 to tokens that belong to annotated evidence for a claim, while assigning a value of 0 to all other tokens. During testing, we compute the average scores for all tokens within a sentence. If the resulting average score exceeds 0.5, we classify the sentence as evidence. Figure \ref{fig:bigbird_example} illustrate this procedure. 

Vanilla Transformer-based architectures, as introduced by \citet{vaswani2017attention}, utilize a complete attention matrix that captures all pairwise interactions among tokens within a given sentence. However, the quadratic time and memory complexity associated with the complete attention matrix imposes limitations on the processing capabilities of these systems, often confining them to moderately long inputs with a typical cutoff point of 512 subwords. Notably, in fact-checking area, a retrieved potential evidence passage can easily surpass this length. For example in CHEF, the average length of an evidence document is 866, which exceeds the maximum length supported by BERT. To overcome this limitation, we turn to BigBird \citep{zaheer2020big}, which incorporates sparse attention patterns. This adaptation enables BigBird to handle sequence lengths of up to 4096 tokens. \citet{zaheer2020big} also demonstrate that BigBird's attention pattern exhibits comparable performance to BERT for short sequence lengths, while outperforming BERT in tasks that involve longer sequences. Figure \ref{fig:bigbird_example} illustrates our framework, highlighting the utilization of BigBird in our approach. 

The models are trained on A100-SXM-80GB GPUs. We fine-tuned the Chinese BigBird\footnote{https://huggingface.co/Lowin/chinese-bigbird-base-4096}. It employs a custom tokenizer, merging jieba tokenizer with BertTokenizer, for processing Chinese text. Optimization leverages the AdamW optimizer (learning rate of 2e-5, epsilon of 1e-8), with no weight decay for specific parameters and a linear learning rate scheduler initiating at a warm-up step count of 0. Training and evaluation phases both utilize a batch size of 16 across 5 epochs.
 
\section{Experiment Setup} \label{exp_app}

In the results presented in Table \ref{tab:pipeline_results}, the translation models initially employ Google/GPT-4 to convert all claims and evidence within the CHEF dataset to English. Subsequently, an English RoBERTa-large is fine-tuned to assess the veracity of these claims using the CHEF training set. For multilingual LLMs, we apply a five-shot in-context learning approach with both GPT-3.5-Turbo and GPT-4-Turbo. Regarding the baseline models—BERT-base, attention-based, and graph-based models—we adhere to the default hyperparameters as delineated in the CHEF study \citep{hu-etal-2022-chef}. We run our experiments on A100-SXM-80GB GPUs. For each pipeline system, we conduct three independent experiments and report the mean values. 

\textbf{Verifiers} We utilize DeBERTa~\citep{he2021deberta} to verify a claim given the selected evidence, using the Chinese version pretrained on the WuDao Corpora~\citep{fengshenbang}. We also compare our results with the baselines in \citet{hu-etal-2022-chef}, including BERT-base~\citep{devlin-etal-2019-bert}, Attention-based ~\citep{gupta-srikumar-2021-x}, and Graph-based~\citep{liu-etal-2020-fine} methods. We also incorporate the RoBERTa-based model~\citep{liu2019roberta}, GPT-3.5-Turbo~\citep{gpt3.5} and GPT-4-Turbo \cite{gpt4} for a more comprehensive comparison. For the GPT models, we use 5 shots for in-context learning..

\section{Comparison of Different Retrievers} \label{compare_retrievers}

Table \ref{tab:retriever_overall} compares the performance of Semantic Ranker and Document-level Retriever. The Document-level Retriever leads to better \textit{Recall@5} and \textit{Marco F1}. \textit{Recall@5} measures the proportion of gold evidence that are successfully retrieved among the top 5 retrieved evidence sentences. 

Although outperforming the Semantic Ranker, the Document-level Retriever only attains a 33.58\% $Recall@5$, indicating the difficulty of evidence retrieval, yet remarkably leads to a 74.46\% Macro F1 score in claim verification. This may be due to the CHEF's gold evidence annotation not being exhaustive, a known issue in datasets with evidence retrieved from the Web~\cite{schlichtkrull2023averitec}, and 
thus the retriever can return correct evidence that was not annotated. Additionally, the model might leverage surface-level patterns in claims to inform verification, which allows for high accuracy even when the available evidence is insufficient.

\begin{table}[h]
    \centering
    \scalebox{0.8}{
    \begin{tabular}{c c c}
    \toprule
        Sentence Retrieval & Recall@5 & Macro F1 \\
        \midrule
        Semantic Ranker & \lpm{21.24}{2.13} &  \lpm{70.58}{1.56}\\
        Document-level Retriever & \blpm{33.58}{2.08}  & \blpm{74.46}{1.78}\\
        \bottomrule
    \end{tabular}
    }
    \caption{Comparison of Semantic Ranker and Document-level Retriever for evidence sentence retrieval with DeBERTa-large.}
    \label{tab:retriever_overall}
\end{table}

\begin{CJK*}{UTF8}{gbsn}
Table \ref{tab:documet_retriever_example} is an example where leveraging document-level information can help with the evidence retrieval. To verify the claim: ``运用红酒含有花青素的原理，可以简单检测红酒的真假。(\textit{The principle that red wine contains anthocyanins allows for a straightforward authenticity test.})", each retriever collects five pieces of evidence. Without additional context, it is not possible to retrieve the sentences highlighted in \textcolor{red}{red} through semantic matching alone. None of these sentences, when considered individually, can be used to verify the claim. However, when taken together, they provide a comprehensive explanation of why anthocyanins can be utilized to test red wine. Having access to the entire document makes it much easier to accurately predict similar examples.

\end{CJK*}

\begin{CJK*}{UTF8}{gbsn}

\begin{table*}[t]
    \centering
    \fontsize{8}{10}\selectfont
    \renewcommand{\arraystretch}{1.5}
    \scalebox{0.88}{
    \begin{tabular}{p{0.35\textwidth} p{0.40\textwidth}}
    \toprule
     \textbf{Semantic Ranker} & \textbf{Document-level Retriever} \\
    \midrule

有一个妙招，一秒钟鉴定红酒真假 \newline (\textit{There's a clever trick, one-second wine authenticity test}) & 
而假红葡萄酒中多由酒精、糖精和香精色素勾兑而成，里面不含花青素 \newline (\textit{Fake red wine is often made by blending alcohol, glycerin, and artificial colorants, without containing anthocyanins}) \\
\midrule
这时，如果红酒变成深蓝色，就是真红酒；如果没有反应，则是假红酒 \newline (\textit{At this point, if the red wine turns deep blue, it's genuine; if there's no reaction, it's fake}) &
\textcolor{red}{由于真正的红葡萄酒中含有丰富的花青素} \newline (\textit{Because authentic red wine contains abundant anthocyanins}) \\
\midrule
如何辨别真假红酒，教你简单一招 \newline (\textit{How to distinguish real from fake red wine, teaching you a simple trick}) & 
\textcolor{red}{花青素在酸性条件下呈现紫红色，而在碱性条件下呈现蓝绿色} \newline (\textit{Anthocyanins appear purplish-red under acidic conditions and bluish-green under alkaline conditions}) \\
\midrule
若是色素勾兑的红酒，颜色则无变化 \newline (\textit{If it's red wine adulterated with colorants, the color remains unchanged}) &
其实，还有一个更简单的方法没说 \newline (\textit{In fact, there's an even simpler method not mentioned}) \\
\midrule
把用水兑开的食用碱水滴在红酒上面； \newline (\textit{Drip food-grade alkali water diluted with water onto the red wine;}) &
如果我们家里的红酒用食用碱检测没有变色，那么基本可以肯定你买到了假酒 \newline (\textit{If our home red wine doesn't change color when tested with food-grade alkali, then it's safe to say you've bought fake wine}) \\

    \bottomrule
    \end{tabular}
    }
    \caption{Evidence sentences retrieved by Semantic Ranker and Document-leverl Retriever for the claim: ``运用红酒含有花青素的原理，可以简单检测红酒的真假(\textit{The principle that red wine contains anthocyanins allows for a straightforward authenticity test.})."}
    \label{tab:documet_retriever_example}
\end{table*}
\end{CJK*}

\begin{figure*}
    \centering
    \includegraphics[width=.85\textwidth]{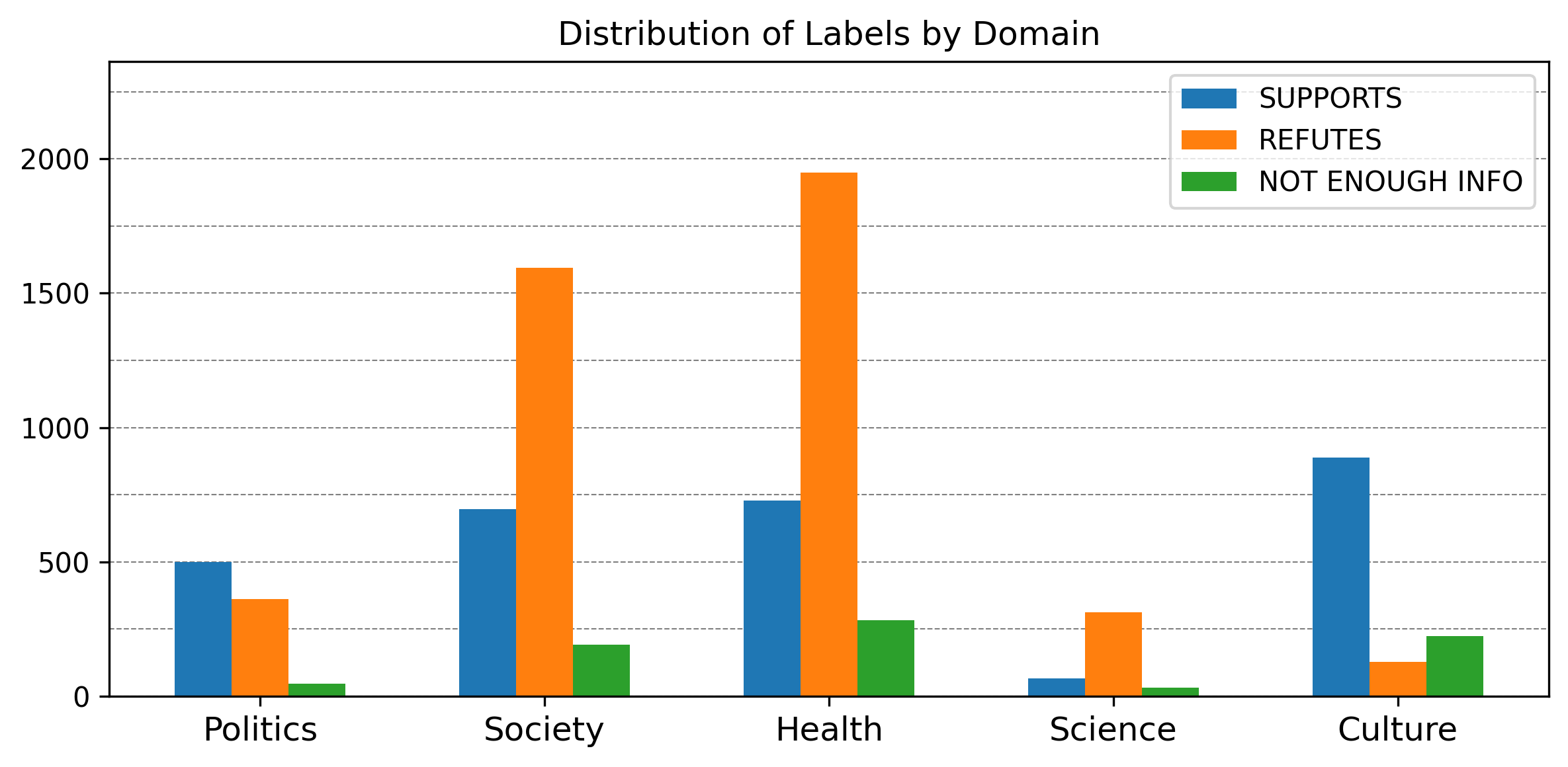}
    \caption{The distribution of labels across different domains in CHEF.}
    \label{fig:distribution}
\end{figure*}

\section{Adversarial Dataset Construction} \label{dataset_construction}

\subsection{Task Definition}

\begin{figure}[h]
    \centering
    \includegraphics[width=0.5\textwidth]{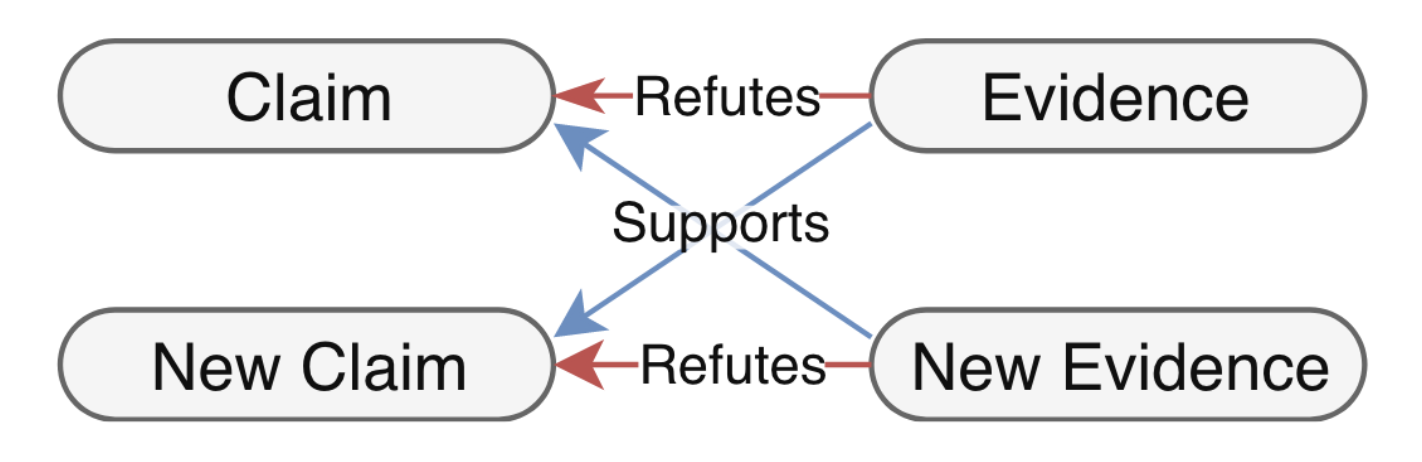}
    \caption{A illustration of the relationship between the original pair and the generated pair \citep{schuster-etal-2019-towards}. }
    \label{fig:adversarial_example}
\end{figure}

To further detect and eliminate bias in CHEF, we propose to generate a new Chinese adversarial dataset for it. We adopt the methodology presented by  \cite{schuster-etal-2019-towards} as our primary framework for constructing a symmetrical dataset for CHEF, as illustrated in Figure \ref{fig:adversarial_example}. Our approach involves generating synthetic claim-evidence pairs that maintain the same relationship (e.g., SUPPORTS or REFUTES) while conveying contrasting factual information. Moreover, we ensure that each sentence in the new pair exhibits the inverse relationship with its corresponding sentence in the original pair. 

\begin{CJK*}{UTF8}{gbsn}

Some new rules have been devised to better suit the Chinese context.
More specifically, when rewriting the given claim ``陈大文在北京称，2020年版第五套人民币5元纸币将发行，防伪性能提升。" (Chen Dawen, announced that the 2020 edition of the fifth series of 5-yuan banknotes will be issued, with improved anti-counterfeiting features, in Beijing.), in our framework, the following rewriting strategies are allowed:

\begin{itemize}
    \item Important nouns that appear in both the claim and the evidence can be modified. These include key information such as time, place, person, and number. Changing these essential terms can alter the original meaning of the sentence. For example, substituting the name ``Chen Dawen" with ``Li Xiaoming," revising the year ``2020" to ``2023," replacing the location ``Beijing" with ``Shanghai," and transforming the denomination ``5 yuan" to ``10 yuan."
    \item Verbs or phrases indicating degrees in both the claim and the evidence can be replaced with their opposites. For instance, substituting ``rise" with ``fall," changing ``increase" to ``decrease," converting ``helpful" to ``unhelpful," replacing ``substantiated" with ``unsubstantiated," and transforming ``no evidence" to ``evidence not found."
\end{itemize}

Note that these methods do not constitute an exhaustive set of legal rewrite methods. They serve as heuristics for the model, which may also employ similar modifications automatically. Similarly, the evidence undergoes a comparable rewriting process. For additional examples of these methods, please refer to Table \ref{tab:adversarial_example}. To rewrite the sentences, we employ the state-of-the-art GPT-4 \citep{gpt4} model ,which has demonstrated human-level performance in various NLP tasks. By leveraging the GPT-4 model, we eliminate the laborious task of human annotation and enhance the diversity of generation through handcrafted rules.

\end{CJK*}

\begin{CJK*}{UTF8}{gbsn}

\begin{table*}[t]
    \centering
    \fontsize{8}{10}\selectfont
    \renewcommand{\arraystretch}{1.5}
    \scalebox{0.88}{
    \begin{tabular}{p{0.12\textwidth} p{0.35\textwidth} p{0.40\textwidth} p{0.10\textwidth} }
    \toprule
    Source & Claim & Evidence & Label \\
    \midrule
    ORIGINAL & 2021年12月31日人民币对美元汇率中间价\textcolor{red}{上}调27个基点。\textit{On December 31, 2021, the central parity rate of the Chinese yuan against the US dollar was increased by 27 basis points.} & 新华社上海12月31日电 来自中国外汇交易中心的数据显示，31日人民币对美元汇率中间价报6.5782，较前一交易日\textcolor{red}{上}调27个基点。\textit{Shanghai, December 31st (Xinhua) - Data from the China Foreign Exchange Trading System showed that the central parity rate of the Chinese yuan against the US dollar was set at 6.5782 on the 31st, representing an increase of 27 basis points compared to the previous trading day.}  & SUPPORT \\
    GENERATED & 2021年12月31日人民币对美元汇率中间价\textcolor{red}{下}调27个基点。\textit{On December 31, 2021, the central parity rate of the Chinese yuan against the US dollar was decreased by 27 basis points.}  & 新华社上海12月31日电 来自中国外汇交易中心的数据显示，31日人民币对美元汇率中间价报6.5782，较前一交易日\textcolor{red}{下}调27个基点。 \textit{Shanghai, December 31st (Xinhua) - Data from the China Foreign Exchange Trading System showed that the central parity rate of the Chinese yuan against the US dollar was set at 6.5782 on the 31st, representing a decrease of 27 basis points compared to the previous trading day.} & SUPPORT \\
    \hline
    ORIGINAL & 奥密克戎对抗体中和作用\textcolor{red}{不存在}逃逸现象。\textit{There is no evidence of escape phenomenon in the neutralizing action of omicron antibodies.} & 结果发现，奥密克戎变异株能被实验中所有单克隆抗体有效中和，\textcolor{red}{没有出现逃逸现象}。\textit{The findings revealed that the omicron variant can be effectively neutralized by all monoclonal antibodies tested in the experiment, with no observed escape phenomenon.} & SUPPORT \\
    GENERATED & 奥密克戎对抗体中和作用\textcolor{red}{存在大量}逃逸现象。 \textit{There is a significant amount of escape phenomenon in the neutralizing action of omicron antibodies.} & 结果发现，奥密克戎变异株能\textcolor{red}{完全抵抗或部分抵抗}实验中所有单克隆抗体的中和作用。 \textit{The results indicate that the omicron variant can completely or partially resist the neutralizing action of all monoclonal antibodies tested in the experiment.} & SUPPORT \\
    \hline
    ORIGINAL & 2020年4月，某男子在公园挖土\textcolor{red}{被警方罚款200元}。 \textit{In April 2020, a man was fined 200 yuan by the police for digging soil in the park. }& 经讯问，邓某承认该微博所述情节均为伪造，其本人并未到过绿博园，更没有被公安机关处罚。\textit{After questioning, Mr Deng admitted that the Weibo post was fabricated, and he had never been to Green Park nor been penalized by the police.} & REFUTE \\
    GENERATED & 2020年4月，某男子在公园挖土被警方制止，但\textcolor{red}{并未罚款}。 \textit{In April 2020, a man was stopped by the police for digging soil in the park but was not fined.} & 据警方透露，某男子于2020年4月在公园非法挖土，发现后被警方罚款200元。\textit{According to the police, the man was found engaging in unauthorized soil excavation in the park in April 2020 and was subsequently fined 200 yuan. }& REFUTE \\
    \hline
    \end{tabular}
    }
    \caption{Examples from the symmetric adversarial dataset are provided to illustrate claim-evidence pairs where the relationship described in the right column is maintained. By combining the generated sentences with the original ones, two additional cases are formed, each with labels that are opposite to one another. The \textcolor{red}{red} texts in Chinese highlight the differences between the claim/evidence before and after the rewrite. }
    \label{tab:adversarial_example}
\end{table*}
\end{CJK*}

\section{Prompt Engineering}

Given the importance of prompt engineering for the quality of the generated data, as well as the scarcity of relevant literature, it is imperative to carefully craft our prompt. To address this challenge, we sought guidance from the empirical findings of the open source community \footnote{https://github.com/f/awesome-chatgpt-prompts}, which provided valuable insights into prompt design practices. Furthermore, we consult the recently published prompt design guideline by \citep{deeplearningai2023prompt} to ensure our approach aligns with the newest recommendations. We conducted extensive experiments to iteratively refine our prompt, culminating in the development of an innovative prompt that not only enhances the quality of generated results but also exhibits versatility, enabling its seamless adaptation to a wide range of tasks.

According to \citet{deeplearningai2023prompt}, the effectiveness of a prompt relies on two key principles. \textbf{Principle 1} emphasizes the significance of providing clear and specific instructions to the model. To achieve this, the prompt should employ delimiters (such as backticks) to clearly demarcate distinct parts of the input. Furthermore, the provision of examples helps the model formulate a ``few-shot" prompt, allowing it to generate responses based on limited examples. \textbf{Principle 2} focuses on optimizing the model's processing by breaking down the full task into several subtasks. This approach guides the model to think step by step, enhancing its performance. The structure of our prompt is outlined in Table \ref{tab:prompt}. 

\begin{CJK*}{UTF8}{gbsn}
\begin{table*}
    \centering
    \fontsize{8}{10}\selectfont
    \renewcommand{\arraystretch}{1.5}
    \scalebox{.95}{
    \begin{tabular}{p{0.5\textwidth} p{0.5\textwidth}}
    \toprule
    \textbf{Explanation of Prompt Design} & \textbf{Prompt Snippet}\\
    \midrule
    Introduce the background of the task and the input format of the data. Define a role for the model. & 我希望你作为一个编辑部的事实核查记者，完成以下的数据标注任务，同时改写声明和证据，使得其各自的含义与原意相反...（I would like you, as an fact-checking journalist, to complete the following annotation task: rewriting claims and evidence so that their respective meanings are the opposite of what they originally meant...）\\
    \midrule
    Give the requirement of how to rewrite the claim. & 第一步：修改声明内容，使得其变成于之前含义相反的内容...(Step 1: Modify the claim to make it have the opposite meaning as before...) \\
    \midrule
    Give the requirement of how to rewrite the evidence accordingly. & 第二步：对应修改后的声明，修改证据的内容...(Step 2: Modify the evidence accordingly, corresponding to the modified claim...) \\
    \midrule
    Give a detailed example and possible rewrite strategies. & 针对例句：[例子]，以下我提供几个理想且合法的修改示例: ...(For the exemplary sentence: [EXAMPLE], I offer the following examples of ideal and legal modifications: ...)\\
    \midrule
    Give a small bunch of human-annotated samples. & 请同时参考以下一些其他例句：示例一；示例二；示例三；...(Please also refer to the following additional example sentences: Example 1; Example 2; Example 3; ...) \\
    \midrule
    Emphasize the key requirement. & 你可以使用上述例子中的修改方式，也可以使用其他修改方法。但是最重要的是要求修改后的证据仍然能支持修改后的声明。(You can use the modification strategies mentioned above as well as other ways to make the changes. However, the most important aspect is to ensure that the modified evidence still supports the modified claim.) \\
    \midrule
    Give the claim and evidence pair that is needed to rewrite delimited by triple backticks. & \`{}\`{}\`{} TEXT \`{}\`{}\`{}\\
    \bottomrule
    \end{tabular}
    }
    \caption{This table outlines the purpose of each snippet in the prompt, explaining the role of each section according to the prompt design principles.}
    \label{tab:prompt}
\end{table*}

\end{CJK*}

\subsection{Quality Control}

Following the data generation process, we generated 250 new claim and evidence pairs. By permuting them under the symmetric setting \citet{schuster-etal-2019-towards}, we obtained an adversarial dataset consisting of 1000 pairs. We then enlisted the participation of two Chinese native speakers to perform annotations on a randomly selected subset of 300 claim-evidence pairs removing their labels, which accounted for 30\% of the total pairs within the symmetric adversarial dataset. These annotations involved assigning one of two labels, namely SUPPORTS, and REFUTES, while also flagging instances of nongrammatical cases. The average agreement between the annotators and the pre-existing dataset labels reached 89\% of the cases, resulting in a Cohen $\kappa$ coefficient of 0.80 \citep{cohen1960coefficient}. It demonstrates that the new claim-evidence pairs generated by GPT-4 mostly remain in their original relation, proving the effectiveness of our method. Additionally, approximately 4\% of the cases exhibited minor grammatical errors or typos. 

\subsection{Error Analysis}

\begin{CJK*}{UTF8}{gbsn}

\begin{table*}[t]
    \centering
    \fontsize{8}{10}\selectfont
    \renewcommand{\arraystretch}{1.5}
    \scalebox{0.88}{
    \begin{tabular}{p{0.12\textwidth} p{0.35\textwidth} p{0.40\textwidth} p{0.10\textwidth} }
    \toprule
    Source & Claim & Evidence & Label \\
    \midrule
    
    ORIGINAL & 2019年1月，成都万象城车祸致\textcolor{red}{一}人死亡。 \textit{In January 2019, a car accident at Chengdu The MixC Mall resulted in one fatality.} & 经交警分局反馈：核实现场\textcolor{red}{无}人死亡，\textcolor{red}{只}有一个伤者。 \textit{According to feedback from the Traffic Police, upon verification, there were no fatalities at the scene, with only one injured individual.} & REFUTE \\
    GENERATED & 2019年1月，成都万象城车祸\textcolor{red}{无}人死亡。\textit{In January 2019, the car accident at Chengdu The MixC Mall resulted in no fatalities.} & 经交警分局反馈：核实现场\textcolor{red}{一}人死亡，\textcolor{red}{还}有一个伤者。\textit{According to feedback from the Traffic Police, upon verification, there was one fatality at the scene, as well as one injured individual.} & REFUTE \\

    \midrule
    ORIGINAL & \textcolor{red}{2018}年春节档总票房累计\textcolor{red}{20.36}亿，“就地过年”让影院更火爆。\textit{During the 2018 Spring Festival season, the total box office revenue reached 2.036 billion RMB, making the cinemas even more popular with the "celebrate the Lunar New Year locally" trend. } & 票房方面，\textcolor{red}{2018}年春节档，中国电影票房\textcolor{red}{20.36}亿，打破\textcolor{red}{2017}年春节档创下的\textcolor{red}{15.06}亿票房纪录，创春节档票房新纪录。\textit{In terms of box office performance, the 2018 Spring Festival season achieved a record-breaking box office revenue of 2.036 billion RMB, surpassing the previous record of 1.506 billion RMB set in the 2017 Spring Festival season and establishing a new record for the Spring Festival box office.}  & SUPPORT \\
    GENERATED & \textcolor{red}{2021}年春节档总票房累计\textcolor{red}{78.45}亿，“就地过年”让影院更火爆。\textit{During the 2021 Spring Festival season, the total box office revenue reached 7.845 billion RMB, making the cinemas even more popular with the "celebrate the Lunar New Year locally" trend.}  & 票房方面，\textcolor{red}{2021}年春节档，中国电影票房\textcolor{red}{78.45}亿，打破\textcolor{red}{2019}年春节档创下的\textcolor{red}{59.06}亿票房纪录，创春节档票房新纪录。 \textit{In terms of box office performance, the 2021 Spring Festival season achieved a record-breaking box office revenue of 7.845 billion RMB, surpassing the previous record of 5.906 billion RMB set in the 2019 Spring Festival season and establishing a new record for the Spring Festival box office.} & SUPPORT \\
    \midrule

    ORIGINAL & 2021年全国有七万硕士在送外卖。 \textit{In 2021, there were 70,000 individuals with master's degrees working as food delivery drivers nationwide.}& 就这样，两个并不具有代表性的“1\%”，被自媒体简单渲染成“全国七万硕士在送外卖”。\textit{Just like that, two non-representative ``1\%"  were sensationalized by the media as ``70,000 master's degree holders nationwide working as food delivery drivers."} & REFUTE \\
    GENERATED & 2021年全国有七万硕士在送外卖\textcolor{red}{为谣言}。 \textit{The claim that there were 70,000 master's degree holders working as food delivery drivers nationwide in 2021 is a rumour.} & 就这样，两个并不具有代表性的``1\%”，得出了一个较为科学的估算，即“全国七万硕士在送外卖”。 \textit{Thus, two non-representative ``1\%" have led to a more scientific estimate of ``70,000 master's degree holders delivering nationwide".}& REFUTE \\
    \midrule
    \end{tabular}
    }
    \caption{Wrongly predicted cases of the DeBERTa-large model after
the inoculation process. The \textcolor{red}{red} texts in Chinese highlight the differences between the claim/evidence before and after the rewrite. }
    \label{tab:error_example}
\end{table*}
\end{CJK*}
After manually examining the wrongly predicted cases for the DeBERTa-large model following the inoculation process, we have identified three primary challenges that current models struggle to address:

\begin{itemize}
    \item Subtle modifications can induce a dramatic change in sentence meaning. In the adversarial CHEF dataset, a large number of statements exhibit slight differences before and after modifications, often differing by only one or two Chinese characters. Given the rich semantic nature of Chinese characters, even a single-word alteration can reverse the entire sentence's meaning. For instance, in the first example of Table \ref{tab:error_example} and the first example of Table \ref{tab:adversarial_example}, minor changes involving a single character completely alter the original meaning. These nuanced distinctions pose difficulties for models to accurately capture. Furthermore, even if these changes are encoded in the model's parameters, they may not receive significant weighting during veracity assessment. 
    
    \item Adversarial CHEF includes numerical reasoning challenges that lack a dedicated mechanism. While the original CHEF dataset contains extensive instances of numbers, there are relatively few statements that necessitate inference from numerical information. In contrast, the adversarial CHEF dataset introduces numerous modifications associated with numbers, requiring the model to determine whether the statements align with the evidence's numerical values. For example, consider the second example in Table \ref{tab:error_example}. However, our current approaches lack a dedicated mechanism to address these numerical issues, resulting in numbers being treated similarly to text.
    
    \item Inferences from implicit or circumstantial evidence present challenges in assessing the claims. In most cases, the evidence is straightforward, enabling easy judgment of the statement's correctness. However, there are instances where the evidence used for inference does not explicitly provide the truth of the statement or directly contradict its content. For instance, the third example in Table \ref{tab:error_example} does not directly specify what is incorrect with the statement (e.g., mentioning that it should be 50,000 instead of 70,000). Instead, the evidence uses terms like ``non-representative" and ``sensationalized" to indirectly point out the unreasonableness of the data results. It is important to note the distinction between this type of challenge and cases involving  ``not enough information," where the former can be deduced through careful inference. Effectively addressing this type of problem requires models with stronger reasoning capabilities. 
    
\end{itemize}

\section{Inoculation by fine-tuning} \label{inoculation}

\begin{figure*}[ht!]
  \centering
  
  \begin{subfigure}{0.5\linewidth}
    \centering
    \includegraphics[width=\linewidth]{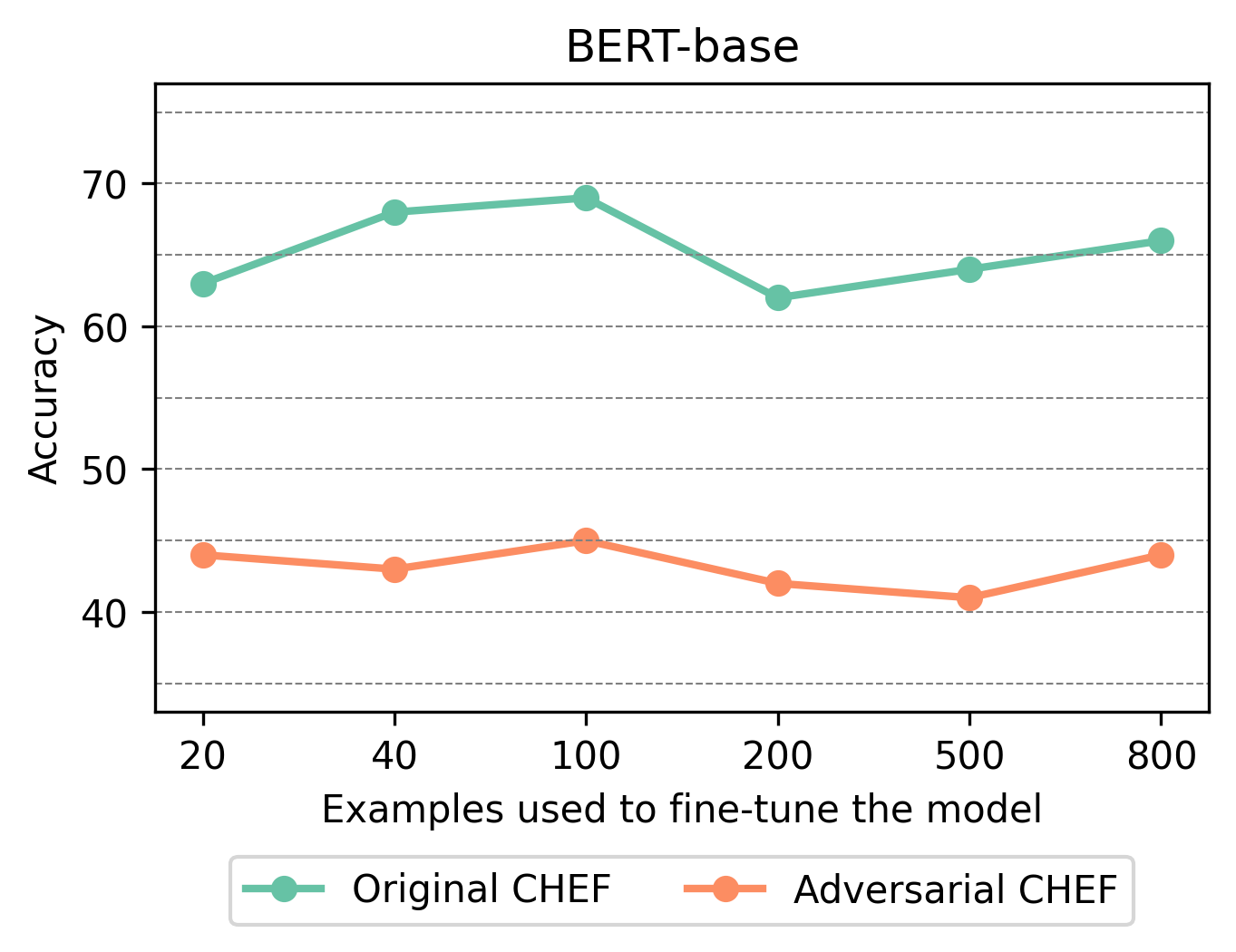}
    \label{fig:sub1}
  \end{subfigure}%
  \hfill
  \begin{subfigure}{0.5\linewidth}
    \centering
    \includegraphics[width=\linewidth]{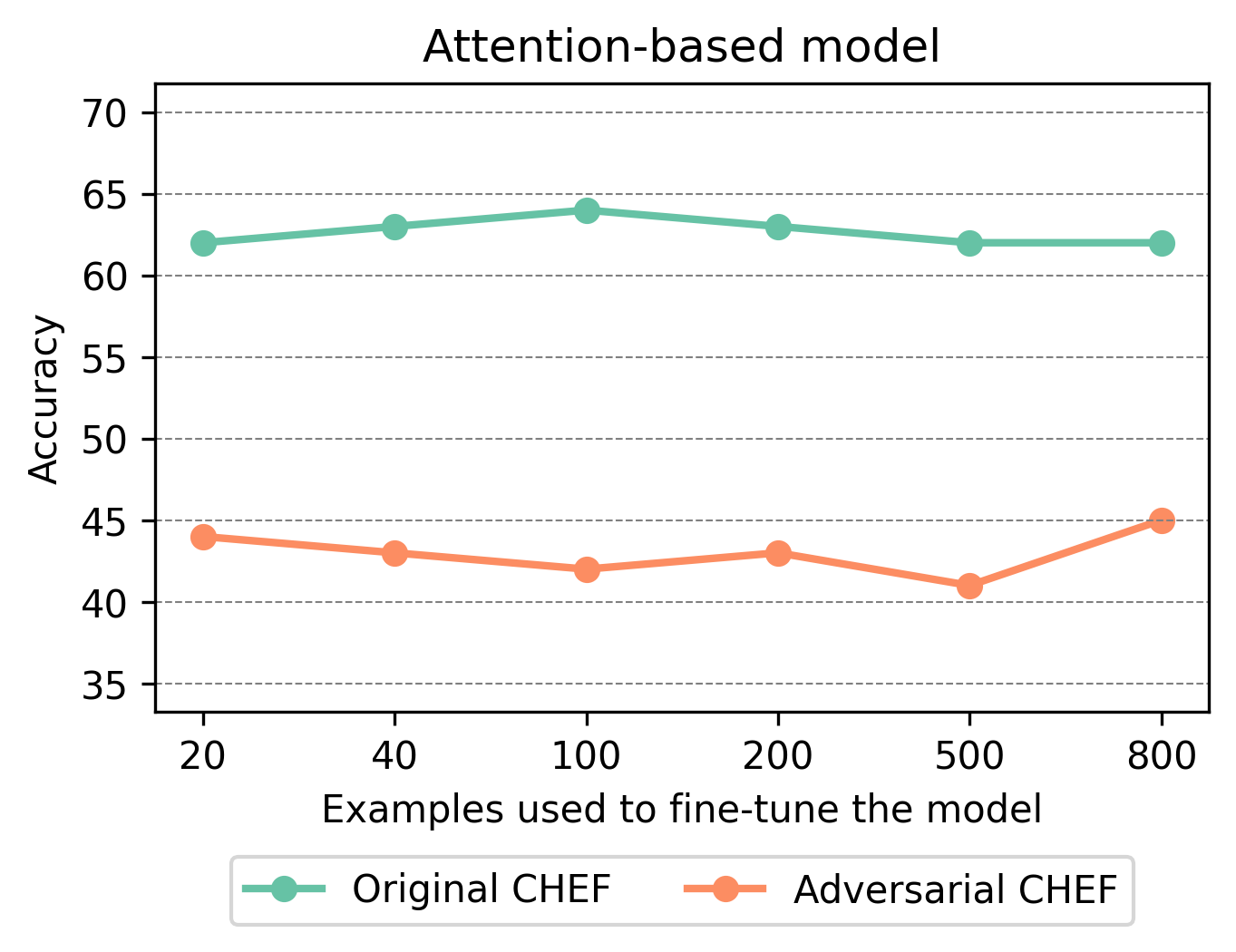}
    \label{fig:sub2}
  \end{subfigure}
  
  \medskip

  \begin{subfigure}{0.5\linewidth}
    \centering
    \includegraphics[width=\linewidth]{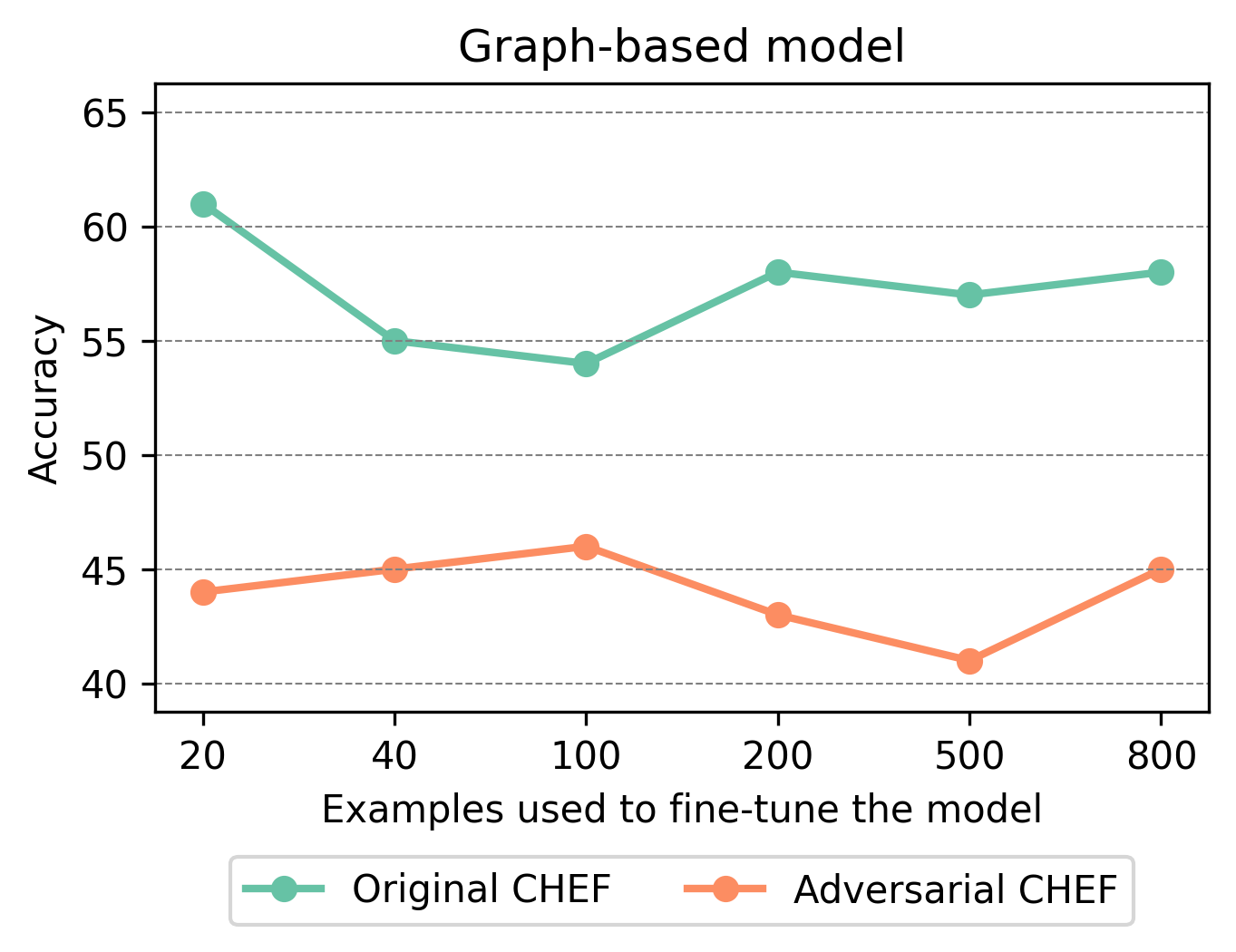}
    \label{fig:sub3}
  \end{subfigure}%
  \hfill
  \begin{subfigure}{0.5\linewidth}
    \centering
    \includegraphics[width=\linewidth]{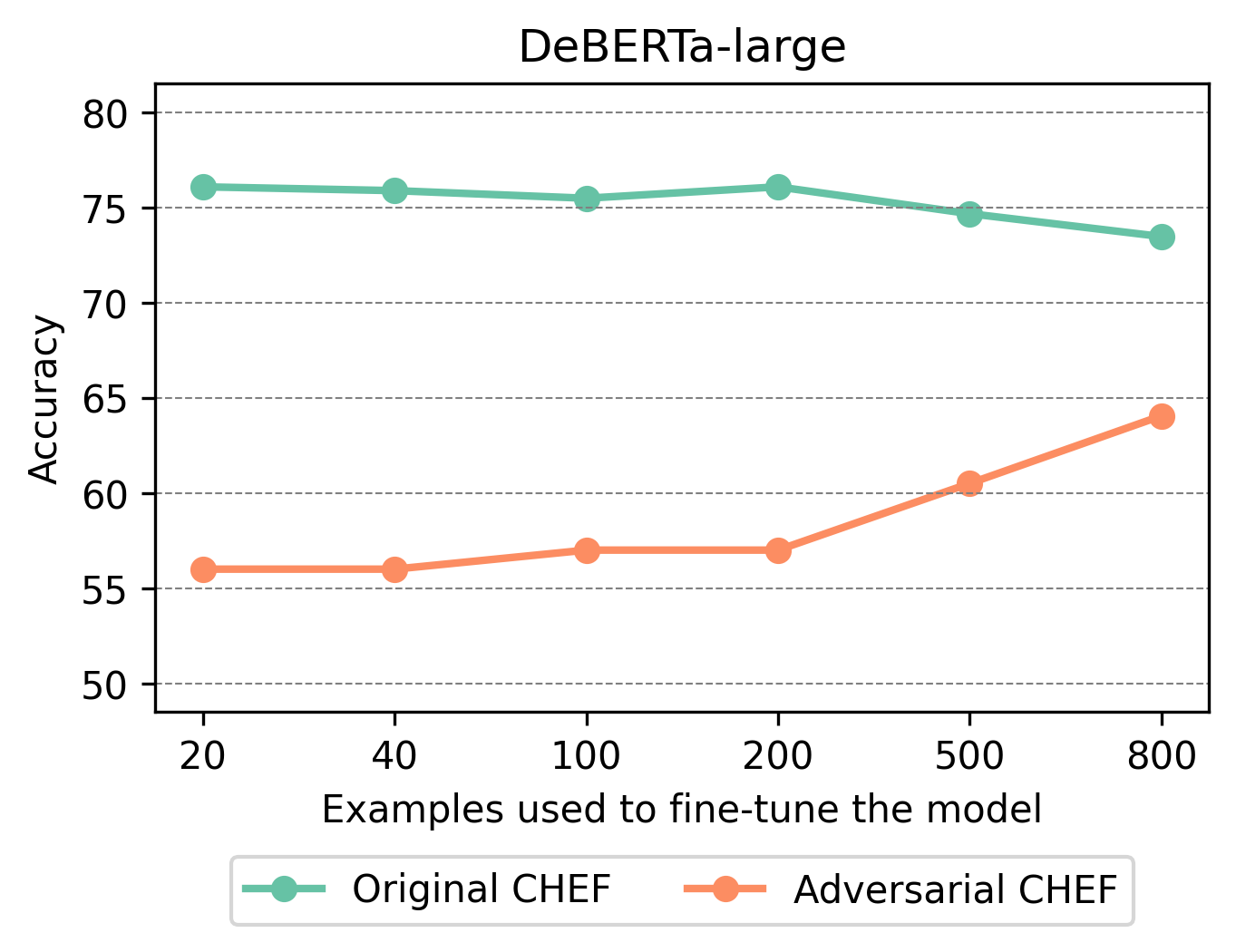}
    \label{fig:sub4}
  \end{subfigure}
  
  \caption{Inoculation results by fine-tuning the model with different sizes of adversarial examples. To evaluate the models, we employ both the original CHEF test set and the adversarial CHEF test set.}
  \label{fig:inoculation_results}
\end{figure*}

Upon evaluating the model with synthetic datasets, it's clear the model underperforms compared to the original benchmarks. The precise weaknesses that these datasets reveal are not immediately revealed. To understand this better, we adopt the method of inoculation by fine-tuning, introduced by \citet{liu-etal-2019-inoculation}. This method allows models to be exposed to a small portion of challenging dataset data to see how the performance changes. 

Post-inoculation, we anticipate three possible outcomes:

\textbf{Outcome 1:} A narrowing of the performance discrepancy between the original and challenge test sets suggests that the challenge data didn't expose model weaknesses but rather a lack of diversity in the original dataset.

\textbf{Outcome 2:} No change in performance on either test set indicates that the challenge dataset has pinpointed a fundamental model flaw, as the model fails to adjust even when familiarized with the challenge data.

\textbf{Outcome 3:} A performance drop on the original test set suggests the fine-tuning skewed the model to suit the challenge data, highlighting a deviation from the original data characteristics. This could be due to differences in label distribution or annotation artifacts that are dataset-specific.

Figure \ref{fig:inoculation_results} shows results from fine-tuning with various amounts of adversarial data. We observe the "performance gap" as the difference in model performance on the original versus adversarial test sets pre-inoculation.

For BERT-base, attention-based, and graph-based models, we observe minor performance changes—Outcome 2—signifying that fine-tuning does not close the performance gap significantly, pointing to a core weakness in adapting to adversarial data distributions.

In contrast, the DeBERTa-large model shows a reduced performance gap post-inoculation, cutting it down by 53\% after fine-tuning with 800 adversarial examples. Its strong performance on the original dataset persists, suggesting DeBERTa's architecture, with its nuanced attention to content, relative, and absolute positions in sentences, equips it to handle slight alterations in claim or evidence more adeptly.

\end{document}